# A Graphical Model Formulation of Collaborative Filtering Neighbourhood Methods with Fast Maximum Entropy Training

**Aaron J. Defazio**  AARON.DEFAZIO@ANU.EDU.AU
**Tibério S. Caetano**  TIBERIO.CAETANO@NICTA.COM.AU
NICTA and Australian National University

## Abstract

Item neighbourhood methods for collaborative filtering learn a weighted graph over the set of items, where each item is connected to those it is most similar to. The prediction of a user's rating on an item is then given by that rating of neighbouring items, weighted by their similarity. This paper presents a new neighbourhood approach which we call *item fields*, whereby an undirected graphical model is formed over the item graph. The resulting prediction rule is a simple generalization of the classical approaches, which takes into account non-local information in the graph, allowing its best results to be obtained when using drastically fewer edges than other neighbourhood approaches. A fast approximate maximum entropy training method based on the Bethe approximation is presented, which uses a simple gradient ascent procedure. When using precomputed sufficient statistics on the Movielens datasets, our method is faster than maximum likelihood approaches by two orders of magnitude.

## 1. Introduction

Recommendation systems have presented new and interesting challenges to the machine learning community. The large scale and variability of data has meant that traditional approaches have not been applicable, particularly those that have quadratic or cubic running time.

This has led to the majority of research taking two tracks: (1) latent-factor models, and (2) neighbourhood models. Latent factor models embed both users and items into a low dimensional space, from which predictions can be computed using linear operations. These include continuous approaches, such as low-rank approximate matrix factorization (Funk, 2006) and binary variable approaches, such as restricted Boltzmann machines (Salakhutdinov et al., 2007).

The second track, neighbourhood models, is the one explored in this work. Neighbourhood methods form a graph structure over either items or users, where edges connect items/users that are deemed similar. Rating predictions are performed under the assumption that users rate similar items similarly (for an item graph) or that similar users have similar preferences (for a user graph) using some form of weighted average (Sarwar et al., 2001).

In this paper we propose a neighbourhood model that treats the item graph as a undirected probabilistic graphical model. This allows us to compute distributions over ratings instead of the point estimates provided by alternative neighbourhood methods. Predictions for a user are performed by simply conditioning the model on her previous ratings, giving a distribution over the set of items she has yet to rate. The predictive rule takes into account non-local information in the item graph, allowing for smaller neighbourhood sizes than are used in other approaches.

We also present an efficient learning algorithm for our model, based on the Bethe entropy approximation. It exploits a decomposition of the variable matrix into diagonal and off-diagonal parts, where gradient ascent need only be performed for the diagonal elements. As our method loops over a set of edges instead of the full set of training data, training is orders of magnitude faster than stochastic gradient descent (SGD) approaches such as Koren (2010). For example, the item graph for the data-set we consider (see Section 6) has approximately 40 thousand edges, which is small compared to the 1 million data-points that are considered in each SGD iteration.

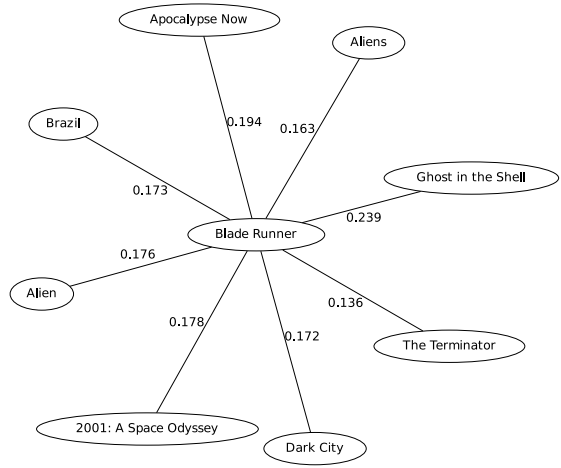

*Figure 1.* Neighbourhood of "Blade Runner" found using the item field method on the MovieLens 1M dataset

## 2. The Item Graph

We begin by introducing the foundations of our model. We are given a set of users and items, along with a set of real valued ratings of items by users. Classical item neighbourhood methods (Sarwar et al., 2001) learn a graph structure over items $i = 0 \ldots N-1$, along with a set of edge weights $s_{ij} \in \mathbb{R}$, so that if a query user $u$ is presented, along with his ratings for the neighbours of item $i$ ($r_{uj}, j \in ne(i)$), the predicted rating of item $i$ is

$$r_{ui} = \mu_i + \frac{\sum_{j \in ne(i)} s_{ji} (r_{uj} - \mu_j)}{\sum_{j \in ne(i)} s_{ji}},$$

where $\mu_i \in \mathbb{R}$ represent average ratings for that item over all users. See Figure 1 for an example of an actual neighbourhood for a movie recommendation system.

In order to use the above method, some principle or learning algorithm is needed to choose the neighbour weights. The earliest methods use the Pearson correlation between the items as the weights. In our notation, the Pearson correlation between two items is defined as

$$s_{ij} = \frac{\sum_u (r_{ui} - \mu_i)(r_{uj} - \mu_j)}{\sqrt{\sum_u (r_{ui} - \mu_i)^2}\sqrt{\sum_u (r_{uj} - \mu_j)^2}}.$$

The set of neighbours of each item is chosen as the $k$ most similar, under the same similarity measure used for prediction. More sophisticated methods were developed for the NetFlix competition, including the work of Bell & Koren (2007), which identified the following problems with the above:

- Bounded similarity scores can not handle deterministic relations;

- Interactions among neighbours are not accounted for, which greatly skews the results;

- The weights $s_{ij}$ cause over-fitting when none of the neighbours provide useful information.

*Learning* the weights $s_{ij}$ under an appropriate model can alleviate all of these problems, and provide superior predictions (Bell & Koren, 2007), with the only disadvantage being the computational time required for training the model. Learning the neighbourhood structure for such a model is not straightforward due to the potentially quadratic number of edges. In this paper we take the approach used by other neighbourhood methods, and assume that the neighbourhood structure is chosen by connecting each item to it's $k$ most similar neighbours, using Pearson correlation as the similarity measure. We denote this undirected edge set $E$. Structure learning is in principle possible in our model, using variants of recently proposed methods for covariance selection (Duchi et al., 2008). Unfortunately such methods become unusable when the number of items considered exceeds a few thousand.

In the next section, we show that the edge weights can be interpreted as parameters of a distribution defined by a particular graphical model. This interpretation leads to fundamentally different rating and training methods, which we explore in Sections 4 and 5 respectively.

## 3. The Item Field Model

Undirected graphical models are a general class of probability distributions defined by a set of feature functions over subsets of variables, where the functions return a local measure of compatibility. In our case the set of variables is simply the set of items, whose values we treat as continuous variables in the range 1 to 5. For any particular user, given the set of their items ratings ($R_K$), we will predict their ratings on the remaining items ($R_U$) by conditioning on this distribution, namely computing expectations over $P(R_U | R_K)$.

The most common feature domains used are simple tuples of variables $(i, j)$, which can be equated with edges in the graphical model, in our case the item graph. We will additionally restrict ourselves to the class of log-linear models, which allows us to write the

general form of the distribution as

$$P(r; \Theta) = \frac{1}{Z(\Theta)} \exp\left[-\sum_{(i,j) \in E} \Theta_{ij} f_{ij}(r_i, r_j)\right],$$

where $\Theta$ is the set of features weights, and $Z$ is the partition function, whose value depends on the parameters $\Theta$, and whose purpose is to ensure the distribution is correctly normalized. The choice of feature function for log-linear models is problem dependent, however in our case two choices stand out. We want functions that encourage smoothness, so that a discrepancy between the ratings of similar items is penalized. We propose the use of squared difference features

$$f_{ij}(r_i, r_j) = \frac{1}{2}((r_i - \mu_i) - (r_j - \mu_j))^2.$$

These features, besides being intuitive, have the advantage that for any choice of parameters $\Theta$ we can form a Gaussian log-linear model that defines that same distribution.

In a Gaussian log-linear model, pairwise features are defined for each edge as $f_{ij}(r_i, r_j) = (r_i - \mu_i)(r_j - \mu_j)$, and unary features as $f_i(r_i) = \frac{1}{2}(r_i - \mu_i)^2$. The pairwise feature weights $\Theta_{ij}$ correspond precisely with the off diagonal entries of the precision matrix (that is, the inverse of the covariance matrix). The unary feature weights $\theta_i$ correspond then to the diagonal elements. Thus we can map the squared difference features to a constrained Gaussian, where the diagonal elements are constrained so that for all $i$,

$$\Theta_{ii} = -\sum_{j \in ne(i)} \Theta_{ji}.$$

We will denote the sparse symmetric matrix of weights for both the Gaussian and squared difference forms as $\Theta$, with the diagonal constrained in this way. This allows us to freely convert between the two forms.

We will impose an additional constraint on the allowable parameters, that each off-diagonal element is non-positive; this constraint ensures that only similarity (as a opposed to dissimilarly) is modelled, and prevents numerical issues in the optimization discussed in Section 5.

## 4. Prediction rule

The feature functions defined in the previous section appear somewhat arbitrary at first. We will now show that they are additionally motivated by a simple link to existing neighbourhood methods. Consider the case of predicting a rating $r_{ui}$, where for user $u$ all ratings $r_{uj}, j \in ne(i)$ are known. These neighbours form the Markov blanket of node $i$. In this case the conditional distribution under the item field model is:

$$\mathcal{N}\left(r_{ui}; \mu_{i|-i}, \frac{1}{\sigma^2} = \sum_{j \in ne(i)} \Theta_{ji}\right), \text{ where}$$

$$\mu_{i|-i} = \mu_i - \frac{\sum_{j \in ne(i)} \Theta_{ji} (r_{uj} - \mu_j)}{\sum_{j \in ne(i)} \Theta_{ji}}.$$

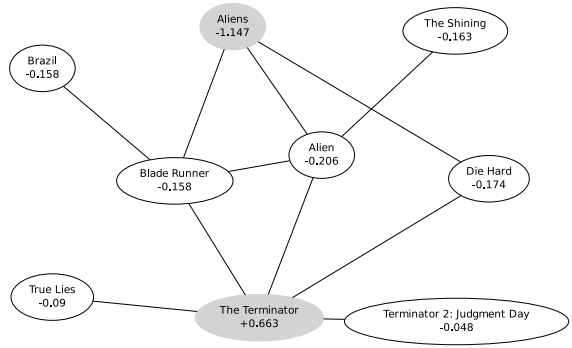

*Figure 2.* Diffusion of ratings, shown as deltas from the item means, for an user that has rated only the shaded items. A hand chosen subgraph of the item graph generated for the 100K movielens dataset is shown.

This is a univariate Gaussian distribution, whose mean is given by a weighted sum of the same form as for traditional neighbourhood methods. In practice we rarely have ratings information for each item's complete neighbourhood, so this special case is just for illustrating the link with existing approaches.

In the general case, conditioning on a set of items $K$ with known ratings $r_K$, with the remaining items denoted $U$, we have:

$$\mathcal{N}\left(r_U; \mu_{U|K}, \Theta_{UU}\right), \text{ where}$$

$$\mu_{U|K} = \mu_U - [\Theta_{UU}]^{-1} \Theta_{UK} (r_K - \mu_K).$$

Thus computing the expected ratings requires nothing more than a few fast sparse matrix operations, including one sparse solve. If the prediction variances are required, both the variances and the expected ratings can be computed using belief propagation, which often requires fewer iterations than the sparse solve operation (Shental et al., 2008).

The linear solve in this prediction rule has the effect of diffusing the known ratings information over the graph structure, in a transitive manner. Figure 2 shows this effect which is sometimes known as spreading activation. Such transitive diffusion has been explored previously for collaborative filtering, in a more ad-hoc fashion (Huang et al., 2004).

Note that this prediction rule is a bulk method, in that for a particular user, it predicts their ratings for all unrated items at once. This is the most common case in real world systems, as all item ratings are required in order to form user interface elements such as top 100 recommendation lists.

## 5. Proposed Training Algorithm: Approximate Maximum Entropy Learning

The traditional way to train an undirected graphical model is using maximum likelihood. When exact inference is used, this is equivalent to maximum entropy training (Koller & Friedman, 2009), as one is the concave dual of the other. However, in practice variational approximations such as the Bethe approximation are used; for example when inference is performed using Belief propagation. In which case the approximate maximum entropy problem is not usually concave, and the solutions of the two are no longer necessarily equivalent.

In discrete models, the (constrained) approximate maximum entropy approach has been shown to learn superior models in some cases, at the expense of training time (Granapathi et al., 2008). In the case of the item field model we will establish that approximate maximum entropy learning is significantly faster, with comparable accuracy.

We first consider the case of a Gaussian model, with the variance-style features described in Section 2, which our constrained Gaussian model builds upon. Let $\Sigma$ denote the empirical rating covariance matrix, which we cap to non-zero only at locations $(i,j) \in E$. The full covariance matrix is dense, however we only need to access the entries at these locations, and it is notationally convenient to treat the rest as zero. The optimization procedure is over the parameters of the beliefs $B = \{b_i, b_{ij}\}$. The approximate maximum entropy objective is subject to the constraints:

$$\forall (i,j) \in E \quad E_{b_{ij}}[x_i x_j] = \Sigma_{ij} \quad (1)$$

$$\forall (i,j) \in V \quad E_{b_i}[x_i^2] = \Sigma_{ii} \quad (2)$$

$$\forall (i,j) \in E \quad \int_{x_i} b_{ij}(x_i, x_j) = b_j(x_j) \quad (3)$$

$$\int_{x_j} b_{ij}(x_i, x_j) = b_i(x_i) \quad (4)$$

$$\forall (i) \in V \quad \int_{x_i} b_i(x_i) = 1 \quad (5)$$

$$\forall (i,j) \in E \quad \int_{x_i} \int_{x_j} b_{ij}(x_i, x_j) = 1. \quad (6)$$

Constraints 3 and 4 are the marginal consistency constraints. When applied to Gaussian beliefs they simply assert that the covariance entries of beliefs with overlapping domains should be equal on that overlap. Constraints 5 and 6 are the normalizability constraints. For Gaussian beliefs they just enforce that the covariance matrices are all positive definite.

In general graphical models, the beliefs are separately parametrized distributions over the same domains as the factors, in our case they take the form of 2D and 1D mean zero Gaussian distributions. We will make use of a representation of the beliefs in a compact form of a single (sparse) symmetric matrix of the same structure as the covariance matrix, which we will denote $C$. This representation simply maps $C_{ij} = E_{b_{ij}}[x_i x_j]$, evaluated at any of the beliefs. This is well defined as the value will be the same for any belief, as noted above. This representation makes the consistency constraints implicit, using what is essentially variable substitution.

The Bethe entropy approximation of the beliefs in our notation is:

$$H_{\text{Bethe}}(C) = \sum_{(i,j) \in E} \log \left( C_{ii} C_{jj} - C_{ij}^2 \right) + \sum_{i \in V} (1 - \deg(i)) \log C_{ii}.$$

Notice that log terms are undefined whenever the belief covariances are not positive definite. Thus the normalizability constraints are also extraneous. So for a purely Gaussian model, the approximate maximum entropy problem simplifies to:

$$\begin{aligned} \underset{C}{\text{maximize}} \quad & H_{\text{Bethe}}(C) \\ \text{s.t.} \quad & C = \Sigma. \end{aligned}$$

Stated this way, the solution is trivial as the constraints directly specify $C$. However, we are interested in learning the weights $\Theta$, which are the Lagrange multipliers of the equality constraints. The Lagrangian is

$$L_\Sigma(C, \Theta) = H_{\text{Bethe}}(C) + \langle \Theta, \Sigma - C \rangle, \quad (7)$$

where $\langle \cdot, \cdot \rangle$ is the standard inner product on matrices. The Lagrangian has derivatives:

**Algorithm 1** Diagonal ascent algorithm for approximate maximum entropy learning

   **input:** covariance $\Sigma$, size $N$, step size $\alpha$
   $C = \Sigma$ and $k = 1$
   **repeat**
     # Compute $\Theta$, needed for each $C_{ii}$ gradient
     **for** $i = 1$ **to** $N$ **do**
       $\Theta_{ii} = \frac{1}{C_{ii}}$
     **end for**
     **for** $(i, j) \in E$ **do**
       **if** $C_{ii} C_{jj} - C_{ij}^2 > 0$ and $C_{ij} > 0$ **then**
         $\Theta_{ij} = \frac{-C_{ij}}{C_{ii}C_{jj} - C_{ij}^2}$
         $\Theta_{ii} \mathrel{+}= \frac{C_{jj}}{C_{ii}C_{jj} - C_{ij}^2} - \frac{1}{C_{ii}}$
       **else**
         $\Theta_{ij} = 0$
       **end if**
     **end for**
     # Take a gradient step on each $C_{ii}$
     **for** $i = 1$ **to** $N$ **do**
       $C_{ii} \mathrel{+}= \frac{\alpha}{\sqrt{k}}(\Theta_{ii} + \sum_{j \in \text{ne}(i)} \Theta_{ij})$
     **end for**
     # Update the off-diagonal elements
     **for** $(i, j) \in E$ **do**
       $C_{ij} = \Sigma_{ij} - \Sigma_{ii} - \Sigma_{jj} + C_{ii} + C_{jj}$
     **end for**
     $k = k + 1$
   **until** sufficient convergence
   **return** $\Theta$

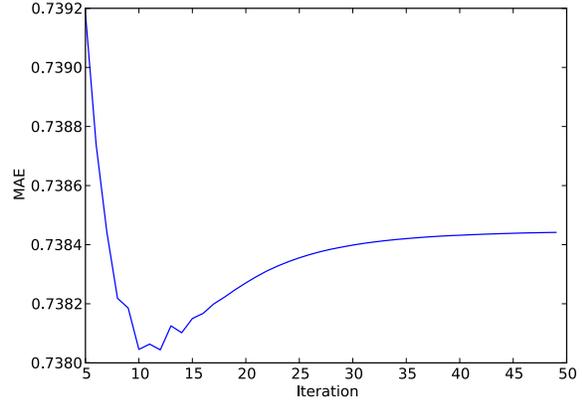

*Figure 3.* Test error as a function of the number of training iterations on the 100K MovieLens dataset. A mild overfitting effect is visible.

$$\frac{\partial L_\Sigma(C, \Theta)}{\partial C} = \frac{\partial H_{\text{Bethe}}(C)}{\partial C} - \Theta$$
$$\frac{\partial L_\Sigma(C, \Theta)}{\partial \Theta} = \Sigma - C.$$

Equating the gradients to zero, gives the following equation for $\Theta$:

$$\Theta = \frac{\partial H_{\text{Bethe}}(C)}{\partial C}|_{C=\Sigma},$$

which gives the closed form solution

$$\begin{aligned}\Theta_{ij} &= \frac{-\Sigma_{ij}}{\Sigma_{ii}\Sigma_{jj} - \Sigma_{ij}^2} \\ \Theta_{ii} &= \frac{1}{\Sigma_{ii}} + \sum_{j \in ne(i)}\left(\frac{\Sigma_{jj}}{\Sigma_{ii}\Sigma_{jj} - \Sigma_{ij}^2} - \frac{1}{\Sigma_{ii}}\right).\end{aligned} \quad (8)$$

The same solution for $\Theta$ can also be obtained using the pseudo-moment matching technique, as $\Sigma$ is a fixed point of the GaBP update equations for this parametrization. If we were applying a vanilla Gaussian model, we could use this result directly. However, for the item field model we have constraints on the diagonal. Using variable substitution on the diagonal, we get the following Lagrangian:

$$\begin{aligned}H_{\text{Bethe}}(C) &+ \sum_{(i,j) \in E} \Theta_{ij}\left(\Sigma_{ij} - \Sigma_{ii} - \Sigma_{jj}\right) \\ &- \sum_{(i,j) \in E} \Theta_{ij}\left(C_{ij} - C_{ii} - C_{jj}\right)\end{aligned}$$

which we denote $L'_\Sigma(C, \Theta)$. It has gradients:

$$\begin{aligned}\frac{\partial L'_\Sigma(C, \Theta)}{\partial C_{ij}} &= \frac{\partial H_{\text{Bethe}}(C)}{\partial C_{ij}} - \Theta_{ij} \\ \frac{\partial L'_\Sigma(C, \Theta)}{\partial C_{ii}} &= \frac{\partial H_{\text{Bethe}}(C)}{\partial C_{ii}} + \sum_{j \in \text{ne}(i)} \Theta_{ij} \\ \frac{\partial L'_\Sigma(C, \Theta)}{\partial \Theta_{ij}} &= \Sigma_{ij} - \Sigma_{ii} - \Sigma_{jj} - C_{ij} + C_{ii} + C_{jj}.\end{aligned}$$

In order to optimize the diagonally constrained objective, we can take advantage of the closed form solution for the simpler unconstrained Gaussian. The procedure we use is given in Algorithm 1. It gives a quality solution in a small number of iterations (see Figure 3). The core idea is that if we fix the diagonal of $C$, the rest of the elements are determined. The gradient of the diagonal elements can be computed directly from $\Theta$, so we recalculate it at each iteration, then take a gradient step. The dual variables are used essentially as notation for the entropy gradients, not in any deeper sense.

### 5.1. Missing Data & Kernel Functions

The training methods proposed take as input a sparse subset of a covariance matrix $\Sigma$, which contains the sufficient statistics required for training. It should be emphasized that we do not assume that the covariance matrix is sparse, rather our training procedure only needs to query the entries at the subset of locations where the precision matrix is assumed to be non-zero.

As our samples are incomplete (we do not know all item ratings for all users), the true covariance matrix is unknown. For our purposes, we form a covariance matrix by assuming the unrated items are rated at their item mean. More sophisticated methods of imputation are possible; we explored an Expectation-Maximization (EM) approach, which did not result in a significant improvement in the predictions made. It did however give better prediction covariances.

In general a kernel matrix can be used in place of the covariance matrix, which would allow the introduction of item meta-data through the kernel function. We left this avenue for future work.

### 5.2. Conditional Random Field Variants

Much recent work in Collaborative filtering has concerned the handling of additional user meta-data, such as age and gender information usually collected by online systems (Stern et al., 2009). These attributes are naturally discrete, and so integrating them as part of the MRF model results in mixed discrete/continuous model. Approximate inference in such as model is no longer a simple linear algebra problem, and convergence becomes an issue. User attributes are better handled in a conditional random field (CRF) model, where the conditional distributions involve the continuous item variables only.

Unfortunately the optimization technique described above does not extend readily to CRF models. Approximate maximum entropy training using Difference-of-convex methods has been applied to CRF training successfully (Granapathi et al., 2008), although such methods are slower than maximum likelihood. We explored CRF extensions using maximum likelihood learning, and while they did give better ratings predictions, training was slow due to the large number of belief propagation calls. While practical if the computation is distributed, the training time was still several hundred times slower than any of the other methods we tested.

### 5.3. Maximum Likelihood Learning with Belief Propagation

An objective proportional to the negative log-likelihood for a Gaussian distribution under the Bethe approximation can be derived from the approximate entropy Lagrangian (Equation 7) using duality theory. First note that the Lagrangian can be split as follows:

$$L_\Sigma(\Theta, C) = H_{\text{Bethe}}(C) + \langle \Theta, \Sigma - C \rangle$$
$$= \langle \Theta, \Sigma \rangle - (\langle \Theta, C \rangle - H_{\text{Bethe}}(C));$$

The dual is then formed by maximizing in terms of $C$:

$$\therefore -\log p(\Sigma; \Theta) \propto \max_C L_\Sigma(\Theta, C) =$$
$$\langle \Theta, \Sigma \rangle - \min_C \left( \langle \Theta, C \rangle - H_{\text{Bethe}}(C) \right);$$

The term inside of the minimization on the right is the Bethe free energy (Yedidia et al., 2000). By equating with the non-approximate likelihood, it can be seen that the log partition function is being approximated as:

$$\log Z(\Theta) = -\min_C \left( \langle \Theta, C \rangle - H_{\text{Bethe}}(C) \right)$$

The value of $\log Z(\Theta)$ and a (locally) minimizing $C$ can be found efficiently using belief propagation (Cseke & Heskes, 2011). For diagonally dominant $\Theta$ belief propagation can be shown to always converge (Weiss & Freeman, 2001). The diagonal constraints as well as the non-positivity constraints on the off diagonal elements of $\Theta$ ensure diagonal dominance in this case.

Maximum likelihood objectives for undirected graphical models are typically optimized using quasi-newton methods, and that is the approach we took here. The diagonal constraints are easily handled by variable substitution, and the non-positivity constraints are simple box constraints. We used the L-BFGS-B algorithm (Zhu et al., 1997) – a quasi-newton method that supports such constraints. The log-partition function $\log Z(\Theta)$ is convex if we are able to exactly solve the inner minimization over $C$, which is not the case in general.

## 6. Experiments

For our comparison we tested on 2 representative datasets. The 1M ratings MovieLens dataset (GroupLens Research) consists of 3952 items and 6040 users. As there is no standard test/training data split for this dataset, we took the approach from Stern et al. (2009), where all data for 90% of the users is used for training, and the remaining users have their ratings split into a 75% training set and 25% test set. The 100K ratings MovieLens dataset involves 1682 items and 943 users. This dataset is distributed with five test/train partitions for cross validation purposes which we made use of.

All reported errors use the mean absolute error (MAE) measure ($\frac{1}{N}\sum_i^N |\mu_i - r_i|$). All 2 datasets consist of ratings on a discrete 1 to 5 star scale. Each method we tested produced real valued predictions, and so some scheme was needed to reduce the predictions to real values in the interval 1 to 5. For our tests the values were simply clamped. Methods that learn a user dependent mapping from the real numbers into this interval have been explored in the literature (Stern et al., 2009).

Table 1. Comparison of a selection of models against the item field model with approximate maximum likelihood, exact maximum likelihood and maximum entropy approaches

| Method | MAE | Precomputation (s) | Training (s) |
| --- | --- | --- | --- |
| 100K MovieLens | | | |
| Maximum Entropy (k=10) | 0.7384 | 18.9 | 0.12 |
| Cosine Neigh. (k=10) | 0.8107 | 18.7 | 0 |
| Bethe Maximum Likelihood (k=10) | 0.7390 | 18.9 | 28 |
| Exact Maximum Likelihood (k=10) | 0.7398 | 18.9 | 99 |
| Least Squares Neighbours (k=10) | 0.7510 | 18.9 | 90 |
| Maximum Entropy (k=50) | 0.7439 | 19 | 1.7 |
| Least Squares Neighbours (k=50) | 0.7340 | 19 | 288 |
| Latent Factor Model (50 factors) | 0.7321 | 0 | 215 |
| 1M MovieLens | | | |
| Maximum Entropy (k=10) | 0.6772 | 514 | 0.64 |
| Cosine Neigh. (k=10) | 0.7421 | 513 | 0 |
| Bethe Maximum Likelihood (k=10) | 0.6767 | 514 | 75 |
| Exact Maximum Likelihood (k=10) | 0.6755 | 514 | 2795 |
| Least Squares Neighbours (k=10) | 0.6826 | 514 | 1369 |
| Maximum Entropy (k=50) | 0.6866 | 517 | 7.17 |
| Least Squares Neighbours (k=50) | 0.6756 | 517 | 4551 |
| Latent Factor Model (50 factors) | 0.6683 | 0 | 4566 |

Table 1 shows the results for the two Movielens datasets. Comparisons are against our own implementation of a classical cosine neighbourhood method (Sarwar et al., 2001); a typical latent factor model (similar to Funk (2006) but with simultaneous stochastic gradient descent for all factors) and the neighbourhood method from Koren (2010) (the version without latent factors), which uses a non-linear least squares objective. All implementations were in python (using cython compilation), so timings are not comparable to fast implementations. We also show results for various methods of training the item field model besides the maximum entropy approach, including exact maximum likelihood training.

The item field model outperforms the other neighbourhood methods when sparse (10 neighbour) models are used. Increasing the neighbourhood size past roughly 10 actually starts to degrade the performance of the item field model: at 50 neighbours using maximum entropy training on the 1M dataset the MAE is 0.6866 vs 0.6772 at 10 neighbours. We found this occurred with the other training methods also. This may be caused by an over-fitting effect as restricting the number of neighbours is a form of regularization.

The latent factor model and the least squares neighbourhood model both use stochastic gradient descent for training. They required looping over the full set of training data each iteration. The maximum entropy method only loops over a sparse item graph each iteration which is why it is roughly two thousand times faster to train. Note that the dataset still has to be processed once to extract the neighbourhood structure and covariance values, the timing of which is indicated in the precomputation column. This is essentially the same for all the neighbourhood methods we compared. In an on-line recommendation system the covariance values can be updated as new ratings stream in, so the precomputation time is amortized. Training time is more crucial as multiple runs from a cold-start with varying regularization are needed to get the best performance (due to local minima).

## 7. Related Work

There has been previous work that applies undirected graphical models in recommendation systems. Salakhutdinov et al. (2007) used a bipartite graphical model, with binary hidden variables forming one part. This is essentially a latent factor model, and due to the hidden variables, requires different and less efficient training methods than those we apply in the present paper. They apply a fully connected bipartite graph, in contrast to the sparse, non-bipartite model we use. Multi-scale conditional random fields models have also been applied to the more general social recommendation task with some success (Xin et al., 2009). Directed graphical models are commonly used as a modelling tool, such as in Salakhutdinov & Mnih (2008). While undirected models can be used in a similar way, the graph structures we apply in this work are far less rigidly structured.

Several papers propose methods of learning weights of a neighbourhood graph (Koren, 2010) (Bell & Koren, 2007), however our model is the first neighbourhood method we are aware of which gives distributions over its predictions. Our model uses non-local, transitive information in the item graph for prediction. Non-local neighbourhood methods have been explored using the concept of spreading activation, typically on the user graph (Griffith et al., 2006) or on a bipartite user-item graph (Lie & Wang, 2009).

## 8. Conclusion

We have presented an undirected graphical model for collaborative filtering that naturally generalizes the prediction rule of previous neighbourhood methods by providing distributions over predictions rather than point estimates. We detailed an efficient training algorithm based on the approximate maximum entropy principle, which after preprocessing takes less than a second to train, and is two orders of magnitude faster than a maximum likelihood approach. Our model has fewer parameters than other comparable models, which is an advantage for interpretability and training.

## Acknowledgements

NICTA is funded by the Australian Government as represented by the Department of Broadband, Communications and the Digital Economy and the Australian Research Council through the ICT Centre of Excellence program.